\documentclass[a4paper, 11pt]{article}
\usepackage[utf8]{inputenc}
\usepackage[english]{babel}
\usepackage[T1]{fontenc}
\usepackage{amsmath,amssymb}
\usepackage[colorinlistoftodos]{todonotes}
\graphicspath{{./images/}}
\usepackage{hyperref}
\usepackage{caption}
\usepackage{subcaption}
\usepackage{todonotes} 
\usepackage{url}            
\usepackage{booktabs}       
\usepackage{amsfonts}       
\usepackage{nicefrac}       
\usepackage{microtype}      
\usepackage{archive}

\usepackage{extarrows}

\newtheorem{thm}{Theorem}[section]

\newtheorem{define}{Definition}[thm]

\title{Meta-Learning and representation learner: A short theoretical note}
\author{Mouad El Bouchattaoui
\\ E-mail: \href{mailto:mouad.el-bouchattaoui@centralesupelec.fr}{mouad.el-bouchattaoui@centralesupelec.fr}
\thanks{Personal e-mail: \href{mailto:mouad.elbouchattaoui@gmail.com}{mouad.elbouchattaoui@gmail.com}}
}
\date{October 2020}
\usepackage[acronym,nomain]{glossaries}
\makeglossaries
\newacronym{utc}{UTC}{Coordinated Universal Time}

\begin{document}
\maketitle
{\color{black} \rule{\linewidth}{0.7 mm} }
\printglossary[type=\acronymtype]

Meta-learning, or "learning to learn," is a subfield of machine learning where the goal is to develop models and algorithms that can learn from various tasks and improve their learning process over time. Unlike traditional machine learning methods focusing on learning a specific task, meta-learning aims to leverage experience from previous tasks to enhance future learning.

This approach is particularly beneficial in scenarios where the available data for a new task is limited, but there exists abundant data from related tasks. By extracting and utilizing the underlying structure and patterns across these tasks, meta-learning algorithms can achieve faster convergence and better performance with fewer data. The following notes are mainly inspired from \cite{vanschoren2018meta}, \cite{baxter2019learning}, and \cite{maurer2005algorithmic}.

\section{Intuition and a Loose Formalism}
In conventional supervised machine learning, we have a dataset \(D = \{ (x_1,y_1), \dots, (x_n,y_n) \}\) on which we train a predictive model \(\hat{y} = f_{\theta}(x)\). The parameters \(\theta\) are chosen by minimizing the loss \(\mathcal{L}(D, \theta, \omega)\), where \(\omega\) encodes what we call the meta-knowledge: choice of the optimizers, the class of functions, estimate of initial parameters, etc. Meta-knowledge can thus be any element of the "how-to-learn" strategy.

In classical training, the optimization is performed for a given dataset of points \(D\) and the meta-knowledge is pre-specified. However, meta-learning consists of learning the "how-to-learn" strategy for a given set of tasks. Loosely speaking, a task can be defined by the result of a dataset and a loss function: \(T = \{D, \mathcal{L}\}\). Meta-learning can thus be formulated as a minimization problem: the search for the best meta-knowledge \(\omega\) that minimizes on average the loss of a model trained on \(D\) for a certain \(\omega\):

$$ \underset{\omega}{\min} \{ \mathbb{E}_{T \sim p(T)}[\mathcal{L}(D, \omega)] \} $$

The average is performed here on a set of tasks on which we define a probability distribution \(p(T)\). In practice, we assume to have access to a set of tasks sampled from \(p(T)\). For example, we can suppose to have \(M\) source tasks to be used in the meta-training: \(D_{src} = \{ (D_{src}^{train}, D_{src}^{val})^{(i)} \mid i = 1, \dots, M \}\).

In the meta-training step, we find the best \(\omega\) such that:

$$ \omega^* =  \underset{\omega}{\arg\max} \{ \log(p(\omega \mid D_{src})) \} $$

Very often, the optimization described in the meta-training step can be described by a bi-level optimization problem:

$$\begin{cases} 
      \omega^* =  \underset{\omega}{\arg\min} \{ \sum_{i=1}^{M}\mathcal{L}^{meta}(\theta^{*(i)}(\omega), \omega, D_{src}^{val(i)}) \} \\
      \theta^{*(i)}(\omega) = \underset{\theta}{\arg\min} \{ \mathcal{L}^{task}(\theta, \omega, D_{src}^{train(i)}) \}
\end{cases}$$

$\mathcal{L}^{meta}$ denotes the "outer objective"; the one used in meta-learning. $\mathcal{L}^{task}$ denotes the inner objective; the one used for the base model learning.

In the meta-testing, we suppose to have \(Q\) target tasks: \(D_{target} = \{ (D_{target}^{train}, D_{target}^{val})^{(i)} \mid i = 1, \dots, Q \}\). During the meta-testing step, each target task will be trained using the meta-knowledge \(\omega^*\). For \(i = 1, \dots, Q\):

$$  \theta^{*(i)} = \underset{\theta}{\arg\max} \{ \log(p(\theta \mid \omega^*, D_{target}^{train(i)})) \} $$

One should mention that transfer learning is not a meta-learning strategy. In transfer learning, the prior can be seen as the result of learning on a source task but without a meta-objective that optimizes the "how-to-learn" strategy. The lack of meta-learning loss is also encountered in domain adaptation and data generalization. On the contrary, we can formulate hyperparameter optimization as a meta-learning problem. This gave birth, for example, to the field of neural architecture search in deep learning. A real challenge in meta-learning is the choice of \(\omega\). In practice, this means defining aspects of the learning strategy to be learned and those to be fixed.

\section{Formal Definition} 
Let \( z \) be a point from the task dataset sampled according to a distribution \( D \) on the data: \( z \sim D \). The distribution thus defines the learning task. A dataset point is generated such that \( z = (x, y) \in Z = X \times Y \) with \( X \) as the input space and \( Y \) as the output space. We define the hypothesis map as \( h: X \longrightarrow W \) (where \( W \) is the action space), whose performance on the data distribution \( D \) will be theoretically assessed by the following risk:

$$ R(h, D) = \underset{z \sim D}{\mathbb{E}}[l(h, z)] $$
$$ R(h, D) = \int_{X \times Y} l(y, h(x)) \, dD(x, y) $$
$$ R(h, D) = \int_{Z} l_{h}(z) \, dD(z) $$

where \( l_{H} = \{ l_h \mid h \in \mathcal{H} \} \), \( l: Y \times W \longrightarrow [0, M] \) is the function that defines the loss function, measuring the similarity/dissimilarity between \( y \) and \( h(x) \). \( \mathcal{H} \) denotes the hypothesis space.

Since a hypothesis depends on the sample, we define a learning algorithm \( A \) that maps a data sample (m-sample) to a hypothesis such that \( A : S = (z_1, \dots, z_m) \sim D^{\otimes m} \longrightarrow A(S) \). The hypothesis \( A(S) \) should perform well on the learning task \( D \) by ensuring a small risk \( R(A(S), D) \). We can use the risk on a sample to theoretically assess the performance of a learning algorithm \( A \) on the task \( D \) by \( \underset{S \sim D^{\otimes m}}{\mathbb{E}}[R(A(S), D)] \). In general, the distribution of \( D \) is unknown; therefore, we prefer to measure the performance of the learning algorithm \( A \) conditioned on a data sample \( S \) by ensuring a statistical guarantee of the type \( \forall D \, \mathbb{P}(R(A(S), D) \leq B(\delta, S)) \geq 1 - \delta \).

In the meta-learning setting, we have a sequence of samples \( \textbf{S} = (S_1, \dots, S_n) \), where each sample originates from a certain task \( S_i \sim D_{i}^{\otimes m} \). In Baxter's formalism, an environment of tasks is given by the probability distribution \( E \) defined on the set of all possible tasks \( \{D_1, D_2, \dots \} \). A task is now a random realization following the distribution probability \( E \). The probability of an m-sample (task sample, not a meta-sample) being generated is \( \mathbb{D}_{E}(S) = \underset{D \sim E}{\mathbb{E}}[D^{\otimes m}(S)] \).

A formal definition of a meta-sample is a sample whose elements are generated independently according to \( \textbf{S} = (S_1, \dots, S_n) \sim \mathbb{D}_{E}^{\otimes n} \). A meta-algorithm \( \textbf{A} \) is thus a mapping from a meta-sample \( \textbf{S} \) to a learning algorithm \( \textbf{A}(\textbf{S}) \). The performance of the learning algorithm \( A \) on the environment \( E \) is measured by the transfer risk: 

$$ \textbf{R}(A, E) = \underset{D \sim E}{\mathbb{E}} \left\{ \underset{S \sim D^{\otimes m}}{\mathbb{E}}[R(A(S), D)] \right\} $$

Very often, the generalization power in the case of the meta-algorithm is given by a control over the generalization error in the following sense:

$$ \forall E \, \text{Proba}(\textbf{R}(\textbf{A}(\textbf{S}), E) \leq B(\delta, \textbf{S})) \geq 1 - \delta $$

One should note that the fact that the tasks are not sampled uniformly — \( E \) is a non-uniform distribution — gives a chance to minimize \( R(A, E) \) as the No-Free-Lunch theorem is not applicable.

A meta-sample is also called an n-m sample and can be written in the form of a matrix of data points:

$$ \textbf{z} = \begin{bmatrix}
z_{11} & \dots & z_{1m} \\
z_{n1} & \dots & z_{nm}
\end{bmatrix} $$

We denote \( Z^{(n,m)} \) as the set of such matrices.

\section{Concept of Representation and Generalization Guarantees}

We now split the hypothesis space \( \mathcal{H} = \{ X \longrightarrow W \} \) into \( \{ X \xlongrightarrow{F} V \xlongrightarrow{G} W \} = G \circ F \). Thus, \( \mathcal{H} \) can be written as \( \mathcal{H} = \{ g \circ f \mid g \in G, f \in F \} \). \( F \) will be called the representation space, and \( f \in F \) will be called the representation. A representation learner is defined as a mapping \( \mathcal{A} : \cup_{n,m \geq 1} Z^{(n,m)} \longrightarrow F \). The idea is to find a good representation \( f \) from information given on the environment \( E \) in the n-m sample \( \textbf{z} \). Therefore, we aim to have a small empirical loss (where \( \langle \cdot \rangle \) refers to the empirical mean and \( z_i \) to the i-th row of \( \textbf{z} \)):

$$ E^{*}_{G}(F, \textbf{z}) = \frac{1}{n} \sum_{i=1}^{n} \underset{g \in G}{\inf} \langle l_{g \circ f} \rangle_{z_i} $$

\( E^{*}_{G}(F, \textbf{z}) \) is an estimate of \( E^{*}_{G}(F, E) = \int_{\mathcal{D}} \inf \langle l_{g \circ f} \rangle \, dE \).  
If \( n \) tasks are sufficient, \( \textbf{D} = (D_1, \dots, D_n) \), we will have \( n \) hypotheses \( \mathbf{g} \circ \Bar{f} = (g_1 \circ f, \dots, g_n \circ f) \) such that \( \mathbf{g} \circ \Bar{f} \in G^n \circ \Bar{F} \). We thus minimize the empirical loss:

$$ \frac{1}{n} \sum_{i=1}^{n} \inf \langle l_{g_i \circ f} \rangle_{z_i} $$

For the sake of good generalization, one should have a bound over \( \mathbb{P}\{ Z \in Z^{(n,m)} \mid d_v(E(\textbf{A}(z), Z), E(\textbf{A}(Z), \textbf{D})) > \alpha \} \) with \( \mathbb{P} = P_{1}^m \otimes \dots \otimes P_{n}^m \) or over \( \mu\{ Z \in Z^{(n,m)} \mid d_v(E^{*}_{G}(\textbf{A}(Z), Z), E^{*}_{G}(\textbf{A}(Z), E)) > \alpha \} \) with \( \mu(S) = \int P_{1}^m \dots P_{n}^m \, dE(P_1, \dots, P_n) \). Here, \( P_1, \dots, P_n \) model the \( n \) tasks.

Let's define \( l_G \) the same way as \( l_H \). Let \( P \) be the probability measure on \( V \times Y \). We define the pseudo-metric:

$$ d_P := \int_{V \times Y} |l_g(v, y) - l_{g'}(v, y)| \, dP(v, y) $$

Let \( \mathcal{N}(\epsilon, l_G, d_P) \) be the size of the smallest \(\epsilon\)-cover of the pseudo-metric space \( (l_g, d_P) \). We define the \(\epsilon\)-capacity as \( C(\epsilon, l_G) = \underset{P}{\mathcal{N}(\epsilon, l_G, d_P)} \). Similarly, we define a pseudo-metric on \( F \) as \( d^{*}_{[P, l_G]}(f, f') = \int_{Z} \sup |l_{g \circ f} - l_{g \circ f'}| \). In this pseudo-metric space, the \(\epsilon\)-capacity will be denoted \( C_{l_G}^{*} = (\epsilon, F) \).
We present the following two theorems \cite{baxter2019learning}. The first theorem bounds the number \( m \) so that we can achieve good generalization given a representation learner averaging over all tasks. The second theorem bounds the number of tasks \( n \) and the number of examples \( m \) per task to achieve good generalization power for a given representation learner and under the same environment. We first introduce some necessary definitions.\\

\begin{define}[Polish space] 
A topological space \( X \) is completely metrizable if there exists a metric \( d \) such that \( (X, d) \) is complete. A separable, completely metrizable space is called a Polish space. 
\end{define}

\begin{define}[Analytic subset] 
An analytic subset of a Polish space can be defined as the continuous image of a Borel set in the Polish space.
\end{define}

\begin{define}[Indexed map]
A map \( \mathcal{H} : Z \longrightarrow [0, M] \) is said to be indexed by the set \( T \) if there exists a function \( f : Z \times T \longrightarrow [0, M] \) such that \( \mathcal{H} = \{ f(., t) \mid t \in T \} \).
\end{define}

\begin{define}[Permissibility]
\( \mathcal{H} \) is permissible if it can be indexed by a set \( T \) which is an analytic subset of a Polish space \( \Bar{T} \), and the indexing function \( f \) is measurable with respect to the product \(\sigma\)-algebra \( \sigma(H) \otimes \sigma(T) \). \( \sigma(T) \) is the Borel \( \sigma \)-algebra induced by the topology on \( T \).
\end{define}

We now extend the concept of permissibility to cover families of hypothesis spaces \( \textbf{H} = \{ \mathcal{H} \} \).

\begin{define}[f-Permissible, Extension]
\( \textbf{H} \) is \( f \)-permissible if there exist sets \( T \) and \( S \) that are analytic subsets of the Polish spaces \( \Bar{T} \) and \( \Bar{S} \), respectively, and a function \( f : Z \times T \times S \longrightarrow [0, M] \) measurable with respect to \( \sigma_{H_{\sigma}} \otimes \sigma(H) \otimes \sigma(T) \), where \( H_{\sigma} = \{ h \mid h \in \mathcal{H} \text{ and } \mathcal{H} \in \textbf{H} \} \), such that:
$$ \textbf{H} = \left\{ \{ f(., t, s) \mid t \in T \}, s \in S \right\} $$
\end{define}

\begin{thm}
Suppose \( F, G \), and \( l \) are such that the family of hypothesis spaces \( \{ l_{G \circ f | f \in F} \} \) is \( f \)-permissible. For all \( 0 < \alpha < 1 \), \( 0 < \delta < 1 \), \( v > 0 \), for any representation learner \( \mathcal{A} \) with values in \( G^n \circ \Bar{F} \), if
$$ m \geq \frac{8M}{\alpha^2 v} \left( \ln(C(\epsilon_1, l_G)) + \frac{1}{n} \ln\left( \frac{4C^{*}_{l_G}(\epsilon_2, F)}{\delta} \right) \right) $$
where \( \epsilon_1 + \epsilon_2 = \frac{\alpha v}{8} \), then
$$ \mathbb{P}\{ Z \in Z^{(n,m)} \mid d_v(E(\mathcal{A}(z), Z), E(\mathcal{A}(Z), \textbf{D})) > \alpha \} \leq \delta $$
\end{thm}

\begin{thm}
Let \( F, G \), and \( l \) be as in the previous theorem. Let \( Z \) be an \( n \)-\( m \) sample drawn according to the environmental measure \( Q \). For \( 0 < \alpha, \delta, \epsilon_1, \epsilon_2 < 1 \), \( v > 0 \), and \( \epsilon_1 + \epsilon_2 = \frac{\alpha v}{16} \), if
$$ n \geq \frac{32M}{\alpha^2} \ln\left( \frac{8C_{l_G}^{*}\left( \frac{\alpha v}{16}, F \right)}{\delta} \right) $$
and
$$ m \geq \frac{32M}{\alpha^2 v} \left[ \ln(C(\epsilon, l_G)) + \frac{1}{n} \ln\left( \frac{8C_{l_G}^{*}(\epsilon_2, F)}{\delta} \right) \right] $$
for any representation learner \( \mathcal{A} : \cup_{n,m \geq 1} Z^{(n,m)} \longrightarrow F \), we have:
$$ \mathbb{P}\{ Z \in Z^{(n,m)} \mid d_v(E_{G}^{*}(\mathcal{A}(Z), Z), E_{G}^{*}(\mathcal{A}(Z), Q)) > \alpha \} \leq \delta $$
\end{thm}

\bibliography{refs.bib}
\bibliographystyle{apalike}
\end{document}